\pdfoutput=1
\documentclass[journal]{IEEEtai}

\newif\ifanon
\anonfalse

\usepackage{graphicx}
\usepackage{amsmath}
\usepackage{amssymb}
\usepackage{booktabs}
\usepackage{multirow}
\usepackage{array}
\usepackage{algorithm}
\usepackage{algorithmic}
\usepackage{xcolor}
\usepackage[hidelinks,breaklinks]{hyperref}
\usepackage{url}
\ifanon\else\usepackage{orcidlink}\fi

\graphicspath{{img/}{./}}
\newcommand{\zenodourl}{\url{https://doi.org/10.5281/zenodo.24279433}}

\begin{document}

\title{Layerwise Learning-Rate Scaling in Vision Networks: Effective-Rate Identification, Power-Mean Aggregation, and Budget Dependence}

\ifanon
\else
  \author{Chandini~Vysyaraju\orcidlink{0009-0000-7167-0798},
          Raghuvir~Duvvuri,
          Radu~Timofte\orcidlink{0000-0002-1478-0402},
          and~Dmitry~Ignatov\orcidlink{0000-0002-6339-1200}%
  \thanks{This work has been submitted to the IEEE for possible publication. Copyright may be transferred without notice, after which this version may no longer be accessible.}%
  \thanks{The authors are with the Computer Vision Lab, CAIDAS \& IFI,
  University of W\"urzburg, 97074 W\"urzburg, Germany
  (e-mail: dmitry.ignatov@uni-wuerzburg.de).}%
  \thanks{Code and data: the layerwise learning-rate functionality of NNGPT is
  at \url{https://github.com/ABrain-One/nn-gpt/tree/main/ab/gpt/brute/llr}. The
  38 strategies, the 936 strategy-injected models and every per-seed record are
  in the NN Dataset, \url{https://github.com/ABrain-One/nn-dataset}, where the
  generated models carry the prefixes \texttt{llr-}, \texttt{llr2-} and
  \texttt{llr3-} (screening sweep), \texttt{llrtn-} (controlled experiments)
  and \texttt{llrood-} (generated-recipe re-check). The scripts that recompute
  each statistic: \zenodourl.}}
  \markboth{Preprint}%
  {Vysyaraju \MakeLowercase{\textit{et al.}}: Layerwise Effective Rate}
\fi

\maketitle
\bstctlcite{IEEEtranBSTCTL:ours}

\begin{abstract}
Layerwise learning-rate scaling assigns different learning rates to different
parts of a network. Although established in fine-tuning pretrained models, its
effects in vision networks trained from scratch are commonly assessed by
training each candidate recipe. We show that, in this setting, the rate-related
component of a layerwise strategy is well summarized by an effective learning
rate computable directly from its multiplier profile. Across matched-control
comparisons, changes in test accuracy closely track changes in training loss
($r=-0.996$), which supports an optimization-progress explanation of the
short-budget rate effect. Increasing the untuned base learning rate tenfold
shifts the rate-matched contrast upward in all six tested architectures
($p=0.016$) and increases the number of architectures in which a published
discriminative schedule, excluded from model fitting, outperforms the uniform
control from two to five of six ($p=0.031$). Within the evaluated strategy
space, a parameter-weighted power mean with exponent $0.35$ ($95\%$ confidence
interval $[0.20,0.50]$) accounts for $42.6\%$ of within-architecture variance,
compared with $36.4\%$ for the arithmetic mean. Ranking candidates by this
scalar without training improves the selected candidate's mean percentile rank
from the $26$th to the $10$th and reduces its mean accuracy shortfall relative
to the best candidate from $3.70$ to $1.12$ points. The rate-related effect
decays as the training budget increases; relative to tuned baselines, the
overall strategy effect falls within the prespecified practical-equivalence
margin in two of three architectures. A separate grouping-related contrast can
persist at long budgets, although its direction and magnitude depend on the
architecture.
\end{abstract}

\begin{IEEEkeywords}
Effective learning rate, layerwise learning rate, statistical evaluation,
training-recipe search, zero-cost proxies.
\end{IEEEkeywords}

\IEEEpeerreviewmaketitle

\section{Introduction}
\IEEEPARstart{N}{eural} architecture search (NAS)~\cite{zoph2017nas,liu2019darts}
automates \emph{what the network is}. A complementary design choice is
\emph{how it is trained}, including layerwise learning-rate (LLR) scaling,
which assigns different rates to different depths. Discriminative fine-tuning
is established for pretrained models~\cite{howard2018ulmfit,zhang2021revisiting,clark2020electra,bao2022beit},
and layerwise adaptive-rate methods such as LARS~\cite{you2017lars} and
LAMB~\cite{you2020lamb} are used in large-batch training. For vision networks
trained \emph{from scratch}, however, the effects of a layerwise schedule and
the way they depend on its multiplier profile remain less well characterized.

This paper develops and evaluates an effective-rate account of layerwise
scaling in this setting. We ask whether the rate-related effect of a strategy
can be summarized by a scalar computed directly from its multiplier profile,
so that candidate recipes can be ranked before each one is trained.
We address this question using matched controls, an intervention on the base
rate, an out-of-sample comparison with a published schedule, and longer-budget
experiments that distinguish rate-related effects from architecture-dependent
grouping effects.

\textbf{Summary of findings.} Across 21 experimental cells, changes in test
accuracy closely track changes in training loss, while the partial association
with the generalization gap is negative once training loss is held fixed
(Sec.~\ref{sec:lossdecomp}), which supports an optimization-progress
interpretation of the rate channel at short budgets. Because a profile and its
outcome may covary for several reasons, we also intervene: raising the base rate
tenfold shifts the pure rate-cut contrast upward in all six tested architectures
(mean shift $+4.7$ points, $p=0.016$) and improves the relative outcome of the
published ULMFiT schedule~\cite{howard2018ulmfit}, which was not in the fitted
strategy space (Secs.~\ref{sec:tworates}--\ref{sec:ulmfit}). The model predicts
the direction of these shifts, not their magnitude.

The parameter-weighted arithmetic mean is a natural baseline summary, but a
concave power mean fits better: profiling gives $p=0.35$, with a
cluster-bootstrap $95\%$ interval of $[0.20,0.50]$ over 22 architecture clusters
(Sec.~\ref{sec:exponent}). The resulting scalar accounts for $42.6\%$ of the
within-architecture variance, compared with $36.4\%$ for the arithmetic mean,
and improves the training-free ranking of the top candidate from the $26$th to
the $10$th percentile (Sec.~\ref{sec:freerank}). The rate-related advantage
diminishes as the training budget grows, and tuned baselines place the strategy
effect within the prespecified equivalence margin for two of the three extended
architectures. A grouping-related contrast at a matched arithmetic mean rate can
persist: it increases on ConvNeXt to $+6.4$ points at 100 epochs but reverses on
GoogLeNet (Sec.~\ref{sec:tuned}). Studies that begin from different base-rate
conditions therefore need not measure the same contribution, which offers one
possible explanation for differing reports on the value of per-group
rates~\cite{mlodozeniec2025completed,zhou2026howtoset}.

\textbf{Contributions.} The study contributes (i) an intervention-supported
identification of the short-budget rate channel
(Secs.~\ref{sec:lossdecomp}--\ref{sec:ulmfit}); (ii) an empirical estimate of
the power-mean aggregation exponent and its uncertainty
(Sec.~\ref{sec:exponent}); (iii) a decomposition of the near-freezing penalty
into effective-rate curvature and a residual threshold-like component
(Sec.~\ref{sec:freeze}); (iv) five practical instruments derived from the
model, including training-free ranking and baseline conversion
(Sec.~\ref{sec:instruments}); and (v) evidence that the tested architecture
descriptors do not predict architecture-specific rate sensitivity
(Sec.~\ref{sec:irreducible}). To our knowledge, this study provides the first joint empirical
identification of the rate channel and measurement of its aggregation exponent
for layerwise strategies in from-scratch vision training.

\section{Background}
\subsection{Layerwise and Adaptive Learning-Rate Scaling}
Discriminative fine-tuning was introduced in ULMFiT~\cite{howard2018ulmfit} and
layerwise decay is now routine for pretrained Transformers in
text~\cite{clark2020electra,zhang2021revisiting} and
images~\cite{bao2022beit}. There the slowed layers hold pretrained features
worth keeping, and transfer studies make the same point from the other side:
slowing early layers helps when useful features already
exist~\cite{yosinski2014transferable} and hinders adaptation when they must
still be learned~\cite{kornblith2019better}. Our rate channel is the
from-scratch counterpart: with no features to protect, what a slowed group
mainly changes is the average step size.

\textbf{Rules that set a multiplier per group.} For networks trained from scratch, the closest precedent is
FreezeOut~\cite{brock2017freezeout}, which gradually reduces each layer's
learning rate to zero before freezing it. FreezeOut studies deliberate
freezing, whereas Sec.~\ref{sec:freeze} quantifies the near-freezing effect
that can arise incidentally when a multiplier is set to $0.01\times$. A method closer to the present setting is TempBalance~\cite{zhou2023tempbalance}, which derives per-layer rates from spectral
statistics of vision networks trained from scratch and maps each layer into
$[0.5\eta,1.5\eta]$ without renormalization. Its mean rate is therefore
not constrained to match that of its control. Two Transformer-oriented
approaches use rates above the base rate. The curvature-based blockwise
schedule~\cite{wang2025sharpness} scales blocks by $10,8,6,4$ while keeping
the normalization blocks at $\eta$. The heavy-tail-guided schedule~\cite{he2026onelr} assigns each layer a rate in $[\eta,5\eta]$, places
the embedding layer at the upper end, and retains the tuned base rate of the
uniform baseline. Layer sampling~\cite{pan2024lisa} argues that the expected
rate should match its dense comparator but applies no correction,
Adam-mini~\cite{zhang2025adammini} reduces each block to one rate by an
arithmetic mean within the block, and LoRA+~\cite{hayou2024lorap}
parameterizes two rates by their \emph{ratio}. Elsewhere per-layer rates are
\emph{derived} rather than searched --- from width
($\mu$P~\cite{yang2021tp5}), from weight-to-gradient norm ratios
(LARS~\cite{you2017lars}, LAMB~\cite{you2020lamb}), or from a modular
norm~\cite{bernstein2025modular} --- while schedule-level interventions act on
all layers at once~\cite{loshchilov2017sgdr,goyal2017accurate}.

\textbf{Open questions addressed here.} Comparisons against a uniform-rate
control can combine changes in the overall rate with effects of the grouping;
the controls of Sec.~\ref{sec:method-controls} separate them. A second question
is how to summarize a multiplier profile: effective-rate arguments associated with linear scaling~\cite{goyal2017accurate} and $\mu$P~\cite{yang2021tp5} are commonly expressed
through an arithmetic mean, while an analysis of fixed rate budgets across
layers constrains their sum~\cite{pang2026balancing}. In addition, weight
decay can cause realized per-layer rates to differ from their nominal values~\cite{kosson2024rotational,fan2025layerwise}, which further motivates
careful rate matching.

\textbf{Differing reports.} Per-module tuning has been reported to yield a
$27\%$ training-time benefit at a scale of $1.3$ billion
parameters~\cite{mlodozeniec2025completed} and a per-layer-group ratio to
improve loss by up to $6\%$~\cite{filatov2025norm}, whereas a greedy search over
four module groups finds $\Delta L\le0.01$ and recommends a single global
rate~\cite{zhou2026howtoset}. The advantage of $\mu$P over standard
parameterization has also been attributed largely to the embedding-layer
rate~\cite{kalra2026embedding}, and a matched-perturbation argument reaches a
similar conclusion for sharpness-aware training~\cite{haas2024mup2}. These
findings motivate distinguishing rate-related effects from those associated with
grouping.

\subsection{Zero-Cost Prediction and Short-Budget Proxies}
Predicting candidate quality without training has been studied extensively
for neural architectures: zero-cost proxies score a network from a single
minibatch~\cite{abdelfattah2021zerocost}, their reliability is examined in
meta-evaluation studies~\cite{lukasik2024evaluation,feng2025zcrobust}, and
proxy reliability in iterative search has been analyzed
formally~\cite{adhikari2026converg}. Comparable tools for ranking training
recipes are less developed; the scalar $M_p$ provides one, and
Sec.~\ref{sec:proxy} evaluates candidate \emph{ordering} and estimation of an
effect's \emph{magnitude} separately. Reporting
recommendations~\cite{lindauer2020best} and controlled re-evaluations of
NAS~\cite{li2019random,white2023nas} motivate the protocol of
Sec.~\ref{sec:design}, which adds cluster-robust inference over shared
baselines, exact permutation tests at small $n$, and equivalence
testing~\cite{lakens2017equivalence}.

\subsection{LLM-Based Recipe Generation}
Language models now propose architectures and training
code~\cite{romera2024funsearch,zheng2023gpt4nas,chen2023evoprompting} and edit
repositories directly~\cite{jimenez2024swebench,yang2024sweagent}, and agentic
systems now search training recipes themselves, reporting the best of many
short trials against a published starting
recipe~\cite{ning2026autoresearch}. The NNGPT/LEMUR platform used in this study~\cite{kochnev2026nngpt,goodarzi2025lemur} hosts generators for
hyperparameters and schedules~\cite{mateen2026lrsched}. We use one such
generator only to obtain search-selected candidates for
Sec.~\ref{sec:zooconvert}, subject to the applicability condition in
Sec.~\ref{sec:design}; the paper's claims about layerwise rates do not
depend on this generator.

\section{Methodology}
\label{sec:model}
\subsection{Layerwise Strategies and the Search Space}
\label{sec:method-strategies}
A layerwise strategy specifies how the parameters are partitioned, the size
of each group, and the learning-rate multiplier assigned to each group. We define 38 strategies
(Supplementary Table~X): 36 split the parameter tensors, in
\texttt{named\_parameters()} order, into 2--6 contiguous depth groups with
split ratios from equal parts to $0.95/0.05$, and two assign tensors to groups
cyclically. The control is \texttt{llr\_uniform}: a single group at $1.0\times$,
injected using the same code scaffold. By multiplier profile, 26 are
\emph{discriminative} in the ULMFiT sense, 8 \emph{inverted} and 2
non-monotone. Every multiplier is $\le1.0$, so the evaluated strategies reduce or preserve
individual group rates relative to the base rate. This defines the one-sided
profile space studied here; strategies that increase selected groups above the
base rate are outside the present experiments.

\subsection{The Rate Channel and Its Power Mean}
\label{sec:method-rateresponse}
\label{sec:model-rate}
A strategy $s$ assigns a fraction $w_g$ of the model parameters to group $g$
and scales that group's learning rate by $m_g\le1$. The effective-rate
hypothesis states that the strategy affects the outcome through a scalar
summary of the profile $(w_g,m_g)$. The choice of summary is an empirical
question. We consider the family of power means:
\begin{equation}
M_p(s) = \bigl(\textstyle\sum_g w_g m_g^{\,p}\bigr)^{1/p},
\quad
M_0(s) = \exp\bigl(\textstyle\sum_g w_g \log m_g\bigr),
\label{eq:powermean}
\end{equation}
indexed by one exponent $p$ and ordered in $p$, so that $M_p\le M_q$ for $p\le
q$. Three special cases provide useful reference points: $p=1$ is the parameter-weighted arithmetic mean
$\bar m=\sum_g w_g m_g$, the factor by which the strategy scales the average
step size and the summary the literature assumes; $p=0$ is the geometric
mean, the natural aggregate if rates combine on a log scale; $p\to-\infty$
approaches $\min_g m_g$, so the aggregate is dominated by the group with the smallest learning-rate multiplier. Across the
38 strategies $\bar m$ ranges from $0.12$ to $0.78$, and $M_p$ for $p<1$ takes smaller values because these power means assign a
larger penalty to a slowed group than its parameter fraction alone would imply.

The literature offers three rival summaries. The arithmetic mean is the usual
implicit one; a second absorbs the profile into one swept scalar times fixed
per-layer-type constants~\cite{everett2024scaling}; a third replaces it by a
norm of the weights~\cite{filatov2025norm,bernstein2025modular}. A power mean of the multipliers can be computed directly from a strategy's
definition, without a forward pass, enabling the training-free instrument in
Sec.~\ref{sec:freerank}. The hypothesis that a single effective-rate scalar
can govern outcomes also has independent support: matching effective rates
collapses loss trajectories in language-model pretraining~\cite{liu2026elr}.

The \emph{rate-response model} states that, within an architecture $a$ and at
a fixed budget, the measured effect of a strategy is governed by $M_p$,
\begin{equation}
\Delta_{a}(s) \;=\; \beta_a\,\bigl(M_p(s) - M^{\,\ast}_a\bigr) \;+\;
\varepsilon ,
\label{eq:model}
\end{equation}
where $\beta_a>0$ is the architecture's \emph{rate sensitivity},
$M^{\,\ast}_a$ is its \emph{break-even} level (the value at which the
fitted strategy effect is zero), and $p$ is shared across architectures. The
model therefore has two architecture-specific coefficients and one
architecture-independent exponent. For
$p<1$, the power mean is concave: reducing a group's rate lowers the summary
more than its parameter fraction alone would imply. A square-root response
of progress to accumulated step size provides one possible motivation for a
concave summary, but the exponent is estimated empirically here rather than
derived from that mechanism.

Equation~(\ref{eq:model}) specifies four hypotheses before analysis. They are
tested in Secs.~\ref{sec:rateresponse}--\ref{sec:tuned} using the
significance tests prespecified in Sec.~\ref{sec:design}.
\textbf{H1} (dose--response): $\beta_a>0$ for every architecture.
\textbf{H2} (optimization, not regularization): the effect is visible in the
training loss, because an effective rate is an optimization quantity.
\textbf{H3} (intervention): raising $\eta_0$ toward the architecture's
optimum lowers $M^{\,\ast}_a$, reducing the penalty associated with rate
cuts.
\textbf{H4} (grouping): any systematic residual after fitting is associated
with the grouping itself. Claims of superiority are evaluated using the tests
specified in Sec.~\ref{sec:design}.

\subsection{The Floor, and How Much of It Is Curvature}
\label{sec:method-freeze}
Equation~(\ref{eq:model}) does not distinguish strategies that have the same
$M_p$ but distribute their rate reductions differently. The deepest group-level
reduction may introduce an additional threshold-like effect. Let
$m_{\min}(s)=\min_g m_g$ denote the smallest group multiplier. The
\emph{near-frozen} indicator
\begin{equation}
f(s) \;=\; \mathbf{1}\!\left[\,m_{\min}(s) \le 0.01\,\right]
\label{eq:freeze}
\end{equation}
equals one for 13 of the 38 strategies. The full model adds one
architecture-independent coefficient,
\begin{equation}
\Delta_{a}(s) \;=\; \beta_a\bigl(M_p(s) - M^{\,\ast}_a\bigr)
\;-\; \gamma\, f(s) \;+\; \varepsilon ,
\label{eq:model2}
\end{equation}
where $\gamma>0$ denotes the additional penalty associated with assigning a
near-zero multiplier to a group, beyond the contribution captured by the
effective-rate summary. At such small multipliers, a group may make little
progress over the evaluated budget, producing a near-freezing cost analogous
to the deliberate freezing studied in FreezeOut~\cite{brock2017freezeout}.

The two terms explain overlapping subsets of the strategy space, which is why
$p$ is estimated rather than fixed. Strategies with a near-zero group are also
those whose $M_p$ decreases most rapidly as $p$ decreases; consequently, an
arithmetic summary can attribute some of its own misspecification to $f$. The
threshold value of $0.01$ in~(\ref{eq:freeze}) is not fitted; it reflects the
available multiplier values. The data therefore locate any residual threshold
only within the interval from $0.01$ to $0.1$.

\subsection{Two Controls That Separate the Channels}
\label{sec:method-controls}
The controls are designed to separate the rate-related contribution from the
contribution associated with grouping. The \emph{rate-matched} control applies the
strategy's parameter-weighted arithmetic mean, $\bar m(s)$, uniformly. It
thereby removes the grouping while matching the strategy's arithmetic mean
rate. The \emph{mean-preserving} variant
retains the groups and relative multipliers while normalizing their
parameter-weighted arithmetic mean to one. Together with the uniform control and the original strategy, these
arms provide the contrasts analyzed in
Secs.~\ref{sec:tworates}--\ref{sec:tuned}.

\section{Experimental Design and Statistical Protocol}
\label{sec:design}
\label{sec:stats}
The experimental program comprises seven components: a one-epoch screening
sweep used to fit the channel and estimate the exponent (837 trials on 25
architectures and three datasets, of which 738 are matched trials on 23
architectures and two datasets); an ablation of training duration and random-seed effects at ten epochs
across six architectures; a $2\times2$ experiment at a second base rate
($\eta_0=0.1$) on the same six architectures, including the ULMFiT test;
mean-preserving strategies on 20 architectures; a 100-epoch replication with
a rate-matched arm on three architectures; tuned 100-epoch baselines for
those same three architectures; and the generated CIFAR-100 candidate population used in
Sec.~\ref{sec:zooconvert}. All runs use \ifanon
\else nn-dataset at commit \texttt{be8657e5}\fi. Supplementary Sec.~XII and Table~S1 provide the
per-experiment details.

\textbf{Trials, injection and clusters.} Each trial injects the strategy into
the architecture's \texttt{train\_setup()}, replacing the optimizer's
\texttt{self.parameters()} with explicit parameter groups. Of the 837 screening
trials with a resolvable baseline, 738 compare against a freshly trained
\texttt{llr\_uniform} run with identical code and hyperparameters and 99
against a stored accuracy; the fit uses these 738 trials, grouped into 23 (architecture, dataset)
baseline clusters. The six ablation architectures span a range of screening outcomes.
GoogLeNet and ConvNeXt were among the highest win rates, whereas DenseNet had no
matched wins among the 38 strategies and serves as a negative control. The
tuning criterion was specified in advance: the tuned control had to
outperform the constant-rate control and come within two points of LEMUR's
Optuna result. Because SGD did not meet this criterion for ConvNeXt, the
ConvNeXt control was tuned with AdamW.

\textbf{An automatically generated candidate population.} The calibration in Sec.~\ref{sec:zooconvert} requires candidates selected by
a search using stored accuracies. We generated these candidates with
Qwen2.5-Coder-7B-Instruct~\cite{hui2024qwen25coder}, fine-tuned with
LoRA~\cite{hu2022lora} on baseline--strategy pairs from the screening sweep
and then applied to 25 CIFAR-100 models from previously unseen families. Naive context matching succeeded on none of the ten held-out pilot targets.
We therefore applied each edit using \emph{AST-based anchor patching} at the
target model's actual \texttt{self.optimizer} node. This approach succeeded
on $9/10$ pilot targets (one-sided Fisher exact $p=6.0\times10^{-5}$) and on all
$25/25$ targets in the main population (Supplementary Sec.~IX).

\textbf{Choice of comparators.} The target of identification is the effect of a layerwise strategy relative
to a uniform-rate strategy. The primary comparator is therefore a matched
\texttt{llr\_uniform} control trained with identical code and
hyperparameters, supplemented by the two constructed controls in
Sec.~\ref{sec:method-controls}, each of which isolates one contribution. A ranking of
published layerwise rules against one another would not identify the channel,
because those rules differ in mean rate and in grouping at the same time. One
published rule~\cite{howard2018ulmfit} is evaluated unmodified as an
out-of-sample test (Sec.~\ref{sec:ulmfit}), and Sec.~\ref{sec:tuned} adds tuned
baselines.

\textbf{Setup.} All controlled runs use CIFAR-10~\cite{krizhevsky2009cifar} at
64\,px with horizontal flips, SGD with momentum $0.9$, a batch size of 64, dropout
$0.2$, and a constant base rate of $0.01$ (or $0.1$ in
Sec.~\ref{sec:tworates}), with no per-architecture tuning; the trainer stops
any run whose accuracy remains at chance level after the first epoch; these
runs are reported rather than replaced.

\textbf{Statistical protocol.} The analysis decisions were specified before
results were reported; Supplementary Sec.~XIII provides the full details.
Trials sharing an architecture--dataset baseline are treated as a cluster.
Win rates are accompanied by a bootstrap over the 23 clusters, a cluster-level
sign test, and per-trial Wilson confidence intervals~\cite{brown2001interval}. Models
(\ref{eq:model})--(\ref{eq:model2}) use architecture fixed effects and
cluster-robust (CR1) standard errors. For arms with $n\le5$, we use exact
permutation tests~\cite{good2005permutation}, with Benjamini--Hochberg
correction~\cite{benjamini1995fdr} within each of 12 prespecified test
families. The mean-preserving experiment is evaluated using a sign test across the 20
architectures. Predicted changes in contrasts across base rates are tested with
a one-sided Wilcoxon signed-rank test on paired differences; within-run
checkpoints are analyzed with a linear mixed model~\cite{bates2015lme4}. An effect is classified as practically negligible only when two one-sided
tests at $\alpha=0.05$ establish that it lies within the prespecified
$\pm1.53$-point margin. This margin is set to the standard deviation across
fifteen replicates of a single constant-rate configuration.

\textbf{Prespecified and exploratory analyses.} The rate-channel tests,
base-rate predictions, grouping-related contrasts, tuned comparison, and proxy and
baseline analyses evaluate hypotheses specified before the data were examined.
The residual floor was identified by examining residuals from the prespecified
rate-response fit. The exponent is estimated by profiling over a grid fixed in advance; the
selected point estimate is nevertheless data-dependent. We report a
cluster-bootstrap interval and leave-one-architecture-out scores, and base the
interpretation on the interval and predictive ordering rather than on the grid
maximum alone.

\section{Results I: The Channel, Its Exponent and Its Floor}
\label{sec:resultsI}
Tables report accuracies and deltas as fractions; the text reports accuracy
points ($0.01$ corresponds to one percentage point).

\subsection{The Rate Channel}
\label{sec:sweep}
\label{sec:rateresponse}
Outcomes vary substantially across the 738 matched trials. Layerwise scaling
outperforms its baseline in 136 cases ($18.4\%$, cluster-bootstrap $95\%$
interval $[9.4,29.1]\%$), while per-architecture win rates range from $0\%$
to $88.6\%$. After multiple-testing correction, 17 of the 22 architectures show a
significant response. This heterogeneity motivates a dose--response analysis
that asks whether a scalar summary of the multiplier profile captures a
systematic component of the effect.

\begin{figure}[t]
\centering
\includegraphics[width=0.99\linewidth]{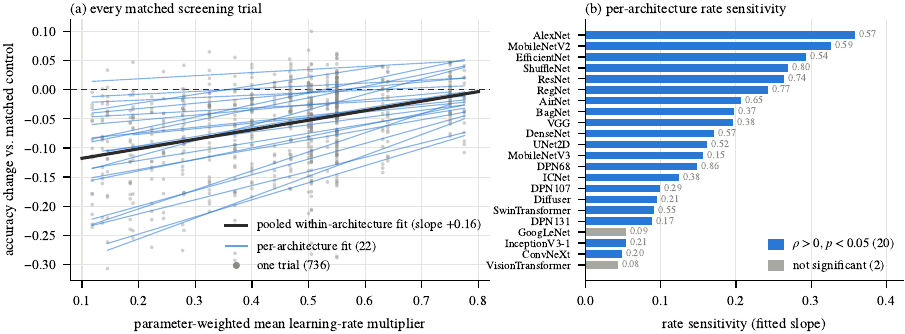}
\caption{The rate channel. (a)~All 736 matched trials against $\bar m=M_1$;
thin lines represent the 22 per-architecture fits of~(\ref{eq:model}),
and the heavy line represents the pooled slope of $+0.163$. (b)~The fitted rate sensitivity $\hat\beta_a$,
annotated by $R^2$; blue bars mark the 20 architectures with $\rho$ significant
at $0.05$.}
\label{fig:rateresponse}
\end{figure}

Fig.~\ref{fig:rateresponse}(a) plots the 736 matched trials of the 22 clusters with at least eight trials against $\bar m$. The
Spearman correlation is \emph{positive for all 22 architectures} (sign test
$p=4.8\times10^{-7}$), individually significant for 20 after
Benjamini--Hochberg correction, with $\rho$ from $+0.11$ (GoogLeNet) to $+0.93$
(ShuffleNet). Fitting~(\ref{eq:model}) at $p=1$ with architecture fixed effects
gives a pooled slope $\hat\beta=+0.163$ (cluster-robust SE $0.020$, $t=8.29$,
$p=4.7\times10^{-8}$), absorbing $36.4\%$ of the within-architecture variance:
a strategy that reduces the mean step size to one-tenth of the baseline
level is associated with an approximately $14.7$-point change in one-epoch
accuracy before its grouping is taken into account. The per-architecture
slope spans a factor of eight, from $+0.043$ (VisionTransformer) to $+0.358$
(AlexNet), with $R^2$ values up to $0.86$ (Fig.~\ref{fig:rateresponse}(b)). The fitted break-even multiplier has median $0.87$ (IQR $[0.58,0.98]$) and exceeds $1.0$ for only four
architectures, suggesting that the untuned base rate is below the fitted
break-even level for most networks. Because this median lies above the maximum
observed arithmetic-mean multiplier of $0.78$, these break-even values are model-based references rather than directly observed optima. The base-rate
intervention in Sec.~\ref{sec:tworates} tests the predicted direction of the
contrast change, not its magnitude.

\subsection{A Concave Power Mean Better Summarizes the Multiplier Profile}
\label{sec:exponent}
\begin{figure}[t]
\centering
\includegraphics[width=0.82\linewidth]{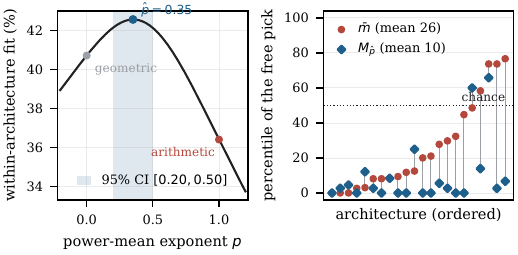}
\caption{The exponent and its effect on candidate selection. (a)~The within-architecture fit
of~(\ref{eq:model}) profiled over $p$, refitting the architecture fixed effects
at every $p$; shading is the cluster-bootstrap $95\%$ interval for the
maximizer. The arithmetic mean accounts for approximately six percentage
points less within-architecture variance than the best-fitting power mean.
(b)~The percentile reached by the single highest-ranked candidate, per
architecture, under $\bar m$ and under $M_{\hat p}$; lower is better, $50$ is
chance.}
\label{fig:exponent}
\end{figure}

Nothing in the effective-rate argument requires $p=1$ in~(\ref{eq:powermean}).
We therefore profile the fit of~(\ref{eq:model}) over $p$ on a prespecified
grid ($-0.20$ to $1.20$ in steps of $0.05$), refitting the architecture fixed
effects at every $p$, on the 736 trials in the 22 clusters of at least eight.
The profile is single-peaked, with its maximum away from $p=1$
(Fig.~\ref{fig:exponent}). The estimated value is
\begin{equation}
\hat p \;=\; 0.35,
\qquad
95\%\ \text{CI}\ [0.20,\,0.50].
\label{eq:phat}
\end{equation}
The confidence interval is obtained from a cluster bootstrap over the 22
architectures ($B=2000$). The cluster-bootstrap distribution yields $\Pr(\hat p<1)=1.000$ and
$\Pr(\hat p>0)=1.000$. Thus, within the
evaluated power-mean family, the estimated aggregation lies between the
arithmetic and geometric means. The slope at the optimum, $+0.167$ (cluster-robust SE $0.019$,
$t=8.59$), is essentially unchanged from the slope at $p=1$. The exponent
therefore changes the relative ordering of strategies without materially
changing the fitted slope.

The power mean improves fit without adding a model parameter. At $\hat p$, a
single scalar accounts for $42.6\%$ of within-architecture variance, compared
with $36.4\%$ for $\bar m$. This recovers six percentage points through the
curvature of the summary, approaching the $43.7\%$ obtained by adding the
separate floor term to the arithmetic model in~(\ref{eq:model2}). The advantage
is also present in leave-one-architecture-out scoring: $\bar m$ reaches $0.353$,
the geometric mean $0.395$, and $M_{\hat p}$, with $p$ refitted on the training
architectures, $0.411$, compared with $0.422$ for the two-term arithmetic
model.

The square-root value $p=0.5$, which is consistent with progress scaling as the
square root of accumulated step size in variance-dominated settings~\cite{wu2026seesaw,shulgin2026hpscaling}, lies at the upper end of the
confidence interval, whereas $p=1$ is outside it. The result is an empirical
measurement of concavity, not a derivation of the exponent from SGD noise
properties. The exponent is estimated primarily on CIFAR-10, which accounts for 20 of
the 22 architecture clusters in the profile. The other two are ImageNette;
adding stored-baseline trials gives three ImageNette clusters, which do not
resolve $p$, and CIFAR-100 contributes no matched trials to
this fit.
Section~\ref{sec:crossds} provides cross-dataset evidence for the positive
rate-response slope, while independent cross-dataset replication of the
exponent remains an open question.

\subsection{The Floor, and How Much of It Was Curvature}
\label{sec:freeze}

\begin{figure}[t]
\centering
\includegraphics[width=0.99\linewidth]{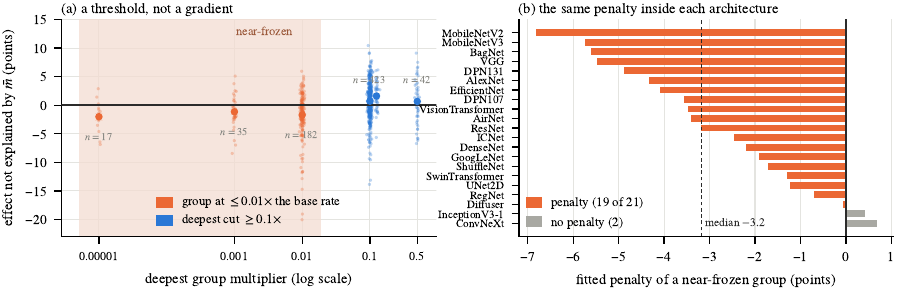}
\caption{The floor. (a)~Residual trial effects not explained by the mean multiplier,
plotted against the multiplier of the most slowed group (log scale). Faint
points represent the 738 trials; heavy markers show means, with error bars
indicating standard errors.
Strategies with a minimum multiplier of $0.01\times$ or less fall below the
rate-response line, whereas those with a minimum multiplier of $0.1\times$
or more lie on or above it; no systematic gradient is apparent within either
region. (b)~The coefficient fitted separately for each architecture; it is
negative in 19 of 21 architectures.}
\label{fig:freeze}
\end{figure}
The residual variation left unexplained by~(\ref{eq:model}) follows a
threshold-like pattern rather than a continuous gradient in severity.
Strategies whose deepest reduction is no more severe than $0.1\times$ lie
\emph{above} the rate-response line by $+0.65$ to $+1.60$ points, whereas
strategies with a group at $0.01\times$ or below lie below it by $-1.15$ to
$-2.02$ points. No further decline is observed down to $10^{-5}$
(Fig.~\ref{fig:freeze}(a)). Fitting the
indicator in~(\ref{eq:freeze}) jointly, as in~(\ref{eq:model2}), gives
$\hat\beta=+0.135$ (SE $0.018$) and $\hat\gamma=2.82$ points (SE $0.43$,
$p=1.2\times10^{-6}$), taking the within-architecture fit from $36.4\%$ to
$43.7\%$ (Fig.~\ref{fig:prediction}(a)). When fitted separately for the 21 architectures with sufficient trials of
both types, the estimated penalty is negative in 19 (median $-3.18$ points;
sign test $p=2.2\times10^{-4}$). The partial rank correlation between
$m_{\min}$ and the residual is positive in $18/21$ architectures
($p=7\times10^{-5}$). The threshold interpretation is also supported by a
diagnostic relevant to the design rule: after excluding strategies with
$m_{\min}\le0.01$, no correlation remains among the other 25 strategies
($p=0.26$).

\textbf{Curvature accounts for much of the apparent floor penalty.}
Re-estimating the model with $M_{\hat p}$ reduces $\hat\gamma$ from
$2.82$ points ($t=-6.60$) to $1.22$ (SE $0.41$, $t=-2.97$). Thus, curvature
in the effective-rate term accounts for $57\%$ of the penalty attributed to
the threshold under the arithmetic summary. The two-term model retains a
within-architecture fit of $43.6\%$. A residual penalty of approximately $1.2$ points remains after this
correction and retains the threshold-like pattern in Fig.~\ref{fig:freeze}(a).
This result motivates a practical precaution within the evaluated strategy
space: avoid assigning multipliers at or below $0.01\times$. Averaging residuals by strategy yields a consistent pattern: after correction,
four of the 38 strategies perform above their predicted values and four perform
below them. All four strategies with below-predicted outcomes contain a group
at or below $0.01\times$ (Supplementary Tables~XXIII--XXV).

\subsection{The Rate-Response Slope Transfers Across Three Datasets}
\label{sec:crossds}
Of the 738 matched trials, 697 are from CIFAR-10. We also include the 99
stored-baseline trials in this analysis. Within each cluster, the stored
baseline is constant, so it shifts all strategy deltas by the same amount and
does not affect within-cluster slopes or rank correlations.
Pooling matched and stored trials within each dataset, the within-cluster slope
of~(\ref{eq:model}) is $+0.165$ on CIFAR-10 (SE $0.019$, $p=2.6\times10^{-8}$,
22 clusters), $+0.133$ on ImageNette (SE $0.047$, $p=0.019$) and $+0.014$ on
CIFAR-100 (SE $0.003$, $p=0.006$), with the same sign across the two baseline types
(Supplementary Table~XXVI). All three dataset-specific slope estimates are positive and statistically
significant. The smaller CIFAR-100 slope is observed on a dataset where one
untuned epoch leaves the controls at only $1$--$19\%$ accuracy, consistent
with lower rate sensitivity as defined in Sec.~\ref{sec:rateresponse}. Per-cluster rank correlations are also positive wherever the cluster size
permits an assessment: all 21 CIFAR-10 clusters and all three ImageNette clusters, and a
positive correlation ($+0.19$) in the one CIFAR-100 cluster large enough for a
rank test. With few independent clusters outside CIFAR-10, this is directional support
rather than a full replication, and the slope magnitudes differ across
datasets.

\subsection{The Tested Architecture Descriptors Do Not Predict Rate Sensitivity}
\label{sec:irreducible}
We tested whether the rate-sensitivity coefficient can be predicted from
commonly available architecture descriptors. Regressing the 21 fitted slopes on parameter count, maximum depth,
layer count, normalization type, the fraction of normalization layers, and the
architecture's uniform-baseline accuracy yields no significant descriptor
association (largest $|r|=0.30$, $p=0.19$). In addition, every candidate
descriptor set has a negative leave-one-architecture-out $R^2$ ($-0.047$ for
baseline accuracy, $-0.181$ for parameter count, and $-0.016$ for all five
together): for these descriptors, the grand mean predicts an unseen
architecture's sensitivity better than the fitted descriptor model.

The eight-fold spread in $\hat\beta_a$ (Fig.~\ref{fig:rateresponse}(b)) is
therefore better measured with a few architecture-specific runs than inferred
from these descriptors; richer architectural or training information may still
have predictive value.

\section{Results II: Identifying the Mechanism}
\label{sec:resultsII}
\subsection{Optimization, Not Regularization}
\label{sec:lossdecomp}
An effective rate is an optimization quantity: it influences how far a run
progresses within a fixed number of steps. An alternative interpretation is that
slowing early layers acts as \emph{regularization}, improving test accuracy
beyond what changes in training loss would predict. The per-seed records separate
these accounts without additional training runs.

For each of the 21 (experiment, architecture, base-rate) cells with per-seed
records at 10 epochs, we calculate the changes in test accuracy, final
training loss, and generalization gap (training minus test accuracy) relative
to that cell's uniform control. Changes in test accuracy are very strongly associated with changes in
training loss: $r=-0.996$ ($p=2.8\times10^{-21}$; Spearman $-0.983$). The
generalization gap is less strongly correlated with the test-accuracy effect
($+0.599$), and its partial correlation with test accuracy is negative after
controlling for training loss ($-0.696$, $p=4.5\times10^{-4}$). This pattern is more consistent with
optimization progress than with an additional benefit attributable to
regularization. Because this decomposition is associational, it is interpreted
together with, rather than as a substitute for, the base-rate intervention.

At 100 epochs, the pattern changes. Training loss is nearly saturated for
all three extended architectures, remaining within $0.005$ of the corresponding
control's value. Across the three cells, the remaining effect is closely
associated with the generalization gap ($r=-0.999$).
With only three cells, this correlation is descriptive rather than a precise
estimate. The optimization-progress component is thus most relevant at shorter
budgets (Sec.~\ref{sec:tuned}).

\subsection{Intervening on the Base Rate}
\label{sec:tworates}
If the strategy acts through an effective-rate channel, increasing $\eta_0$
toward an architecture's preferred operating region should lower its fitted
break-even level. A rate cut that is initially harmful should therefore become
less harmful or beneficial. This directional prediction was specified in
advance.

\begin{figure}[t]
\centering
\includegraphics[width=0.86\linewidth]{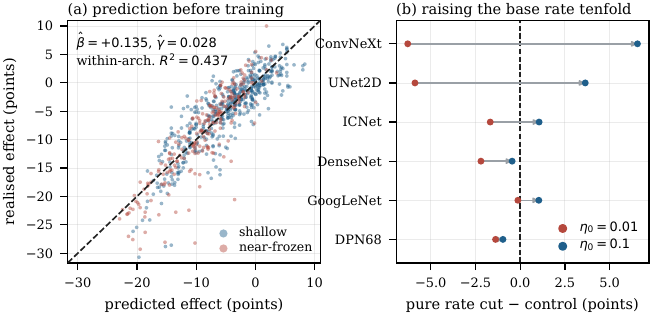}
\caption{The model in use. (a)~Realized versus predicted effects for every
matched trial under~(\ref{eq:model2}) with architecture fixed effects
($\hat\beta=+0.135$, $\hat\gamma=2.82$ points, $R^2=0.437$); color indicates
whether the strategy crosses the floor of~(\ref{eq:freeze}).
(b)~Identification: the pure rate-cut contrast at $\eta_0=0.01$ and
$\eta_0=0.1$ (epoch 10, five seeds), moving upward in $6/6$ architectures.}
\label{fig:prediction}
\end{figure}

Supplementary Table~S2 reports the full $2\times2$ experiment at epoch 10,
with five seeds per arm; Fig.~\ref{fig:prediction}(b) shows the key contrast. At
$\eta_0=0.01$ the pure rate cut --- the rate-matched control --- is below the
uniform control in $6/6$ architectures (mean $-2.9$ points); at $\eta_0=0.1$, it
is above it in $4/6$ (mean $+1.8$). In paired comparisons within each architecture, the pure rate-cut contrast
moves upward in all six architectures (mean shift $+4.7$ points; one-sided
Wilcoxon signed-rank $p=0.016$). The full strategy-minus-control effect also moves upward
in five of the six architectures ($p=0.031$). ConvNeXt moves
furthest: its selected strategy is $3.3$ points \emph{below} its control at
$\eta_0=0.01$ and $17.6$ \emph{above} it at $\eta_0=0.1$ (exact $p=0.008$), and
its break-even level of $0.37$ is among the three lowest of the six. The \emph{size} of the shift is not predicted ($\rho=-0.09$);
we claim the direction, not the magnitude.

\subsection{Out-of-Sample Evaluation of a Published Rule}
\label{sec:ulmfit}
The strategies above were selected by our screening procedure. A more
stringent out-of-sample test therefore evaluates a published rule that was not
used for model fitting. ULMFiT's discriminative schedule~\cite{howard2018ulmfit} provides such a test. Published before this study
and designed for fine-tuning language models, it was not part of the fitted
strategy space. Its profile has $\bar m=0.40$ and $m_{\min}=0.057$, so it
introduces a substantial rate reduction without crossing the near-freezing
threshold. The model therefore predicts that it should behave like a moderate
rate cut and improve as $\eta_0$ increases.

The observed results follow this directional prediction. Applied unchanged
to the same six architectures, ULMFiT outperforms the uniform control in
$2/6$ architectures at $\eta_0=0.01$ and $5/6$ at $\eta_0=0.1$. The
within-architecture effect increases in five of the six architectures
(one-sided Wilcoxon signed-rank $p=0.031$; mean shift $+13.0$ points). On ConvNeXt, the
effect increases from $+5.0$ to $+10.0$ points (exact $q\le0.027$). The
model predicts the direction of this shift and does not calibrate its magnitude.

\subsection{What the Grouping Adds at a Matched Mean Rate}
\label{sec:structure}
The rate-matched control removes grouping while holding the
parameter-weighted arithmetic mean, $\bar m$, fixed. The contrast between
the original strategy and this control therefore measures the contribution
associated with grouping at a matched arithmetic mean rate. Across the 12 (architecture, base-rate) cells with both arms, this contrast
is positive in 10 (median $+0.85$ points; sign test $p=0.039$). Neither negative cell uses a strategy that crosses the near-freezing threshold
in (\ref{eq:freeze}), whereas the largest positive contrasts (ConvNeXt) come from
one that does. Within this comparison, concentrating a rate reduction in groups
often performs better than distributing the same mean reduction uniformly.
The next subsection examines whether this contrast persists at longer budgets.

\subsection{Two Channels, Two Timescales}
\label{sec:tuned}
\label{sec:reversal}
\label{sec:ablation}
The model predicts that the rate-related advantage should diminish as the base
recipe is tuned and as additional training allows optimization progress to
accumulate.

\begin{figure}[t]
\centering
\includegraphics[width=0.90\linewidth]{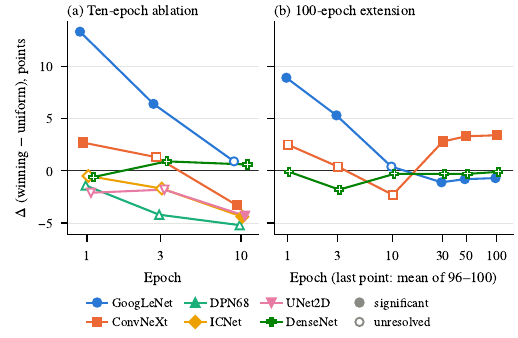}
\caption{Selected strategy minus \texttt{llr\_uniform} under the untuned
constant-rate recipe at base rate $0.01$, in accuracy points, $n=5$ seeds per
arm. (a)~Ten-epoch ablation, six architectures; filled markers are significant
after correction over the 18 cells. (b)~100-epoch extension, three architectures
(last point: mean of epochs 96--100); filled markers have exact permutation
$p<0.05$.}
\label{fig:ablation}
\end{figure}

\textbf{The same recipe, run longer.} Under the untuned constant rate, the
strategy effects decay with budget, with an estimated change of $-0.0197$ per
log-epoch in the mixed model
($95\%$ CI $[-0.0274,-0.0121]$, $p=4.5\times10^{-7}$) --- and the decline is associated with the training budget rather than with
selection: in the unplanned replication, the \emph{unselected} strategies
decline more than the selected ones. At 100 epochs
the three extended architectures show different responses. GoogLeNet's $+8.9$-point effect at epoch 1
decreases to $-0.7$ over epochs 96--100 (exact $p=0.016$, within the equivalence
margin), while DenseNet's effect is $-0.1$ ($p=0.82$). ConvNeXt moves from $2.3$ points below the
control at epoch 10 to $3.4$ points above it at convergence ($p=0.008$)
(Fig.~\ref{fig:ablation}).

\textbf{The baseline itself.} In a one-seed rate sweep (Supplementary Table~XIV), tuning raised the control's
converged accuracy to $90.6\%$ for GoogLeNet, $91.1\%$ for DenseNet and, with
AdamW, $83.2\%$ for ConvNeXt. All three
are released as improved reference baselines. Against these tuned controls (Supplementary Table~XV), the strategy effects
for GoogLeNet ($-0.19$ points, $q=0.048$) and DenseNet ($-0.49$ points,
$q=0.012$) are statistically equivalent to zero within the prespecified
$\pm1.53$-point margin. Both $90\%$ confidence intervals lie entirely
within this margin (TOST $p<10^{-4}$). For ConvNeXt, the strategy is $2.37$ points below the tuned control
($q=0.012$). Thus, its $3.4$-point gain relative to the untuned control does
not persist when the comparison is made against the tuned baseline.

\textbf{The rate-related effect decays; the grouping-related contrast is
architecture-dependent.} For the rate-related effect, fitting
$|\Delta(T)|\propto T^{-b}$ over epochs 2--100 gives $b=+0.20$ for GoogLeNet
and $b=+0.57$ for DenseNet. The tuned-baseline comparison places the remaining
strategy effects within the prespecified practical-equivalence margin for two
of the three architectures. The rate-matched arm provides a separate estimate of the
grouping-related contrast. At a matched arithmetic mean rate, the layerwise
strategy exceeds the rate-matched uniform control by $+4.93$ points
at 10 epochs, $+6.07$ at 50, and $+6.39$ at 100 on ConvNeXt (exact permutation
$p=0.008$ at each budget). On GoogLeNet, the corresponding contrast is $+0.74$
at 10 epochs ($p=0.040$) and reverses to $-0.88$ at 100 ($p=0.008$). Thus,
the grouping-related contrast does not follow a common decay pattern in these two
architectures: it increases on ConvNeXt and changes sign on GoogLeNet. For ConvNeXt, the two results are not contradictory: the strategy is
$2.37$ points below the tuned uniform control but $6.39$ points above the
rate-matched control at 100 epochs because the two comparisons isolate
different contributions.

The practical implication is that searching the mean rate is most useful at
shorter budgets and before the base rate has been tuned, whereas the
grouping-related contrast may remain relevant at longer budgets for some
architectures. Training-algorithm benchmarks often compare time-to-target against
a fixed baseline~\cite{dahl2023benchmarking,kasimbeg2025algoperf}, and retuning
can change method rankings as advantages diminish with scale or
budget~\cite{wen2025fantastic,shen2026looped} and affect conclusions in
reinforcement learning~\cite{eimer2023hyperparameters}; our results provide a
measured example for layerwise strategies.

\section{Results III: Five Practical Uses of the Effective-Rate Model}
\label{sec:resultsIII}
\label{sec:instruments}
\subsection{Ranking Candidates and Predicting Effect Signs Without Training}
\label{sec:freerank}
\label{sec:ranking}
A power mean of a strategy's multipliers can be computed directly from its
definition, allowing a candidate pool to be ranked before training. The
resulting order agrees with the realized ordering: the per-architecture
Spearman correlation is positive in all 22 architectures (median $+0.71$),
consistent with the dose--response relationship in
Sec.~\ref{sec:rateresponse}. The main benefit of the estimated exponent is improved selection of the top
candidate. We take the single highest-ranked candidate, as a practitioner
would when selecting a recipe without additional training, and determine its position in the
realized ordering for each architecture. The arithmetic mean, $\bar m$,
selects a candidate at the $26$th percentile, compared with the $50$th
percentile expected by chance; $M_{\hat p}$ selects one at the $10$th
percentile. The accuracy regret relative to the best candidate decreases from $3.70$ to
$1.12$ points. The zero-cost selection falls within the realized top decile in 17 of 22
architectures under $M_{\hat p}$, compared with 9 under $\bar m$ (Fig.~\ref{fig:exponent}(b)). This gain
comes from reordering: the two summaries have similar fitted slopes but
differ in the penalty they assign to near-frozen groups, which the arithmetic
mean tends to underweight.

When used to predict whether a candidate will outperform its matched uniform
baseline, the same scalar acts as a classifier. Its within-architecture AUC is
$0.858$ under $M_{\hat p}$, compared with $0.837$ under $\bar m$, across the
15 architectures whose trials include both outcomes. The scalar therefore
provides a zero-cost screening score, not an estimate of the magnitude of a
candidate's gain.

\subsection{What Transfers Across Architectures}
\label{sec:transfer}
Recipe effects show partial transfer across architectures. Across the 190
pairs of CIFAR-10 architectures, the average rank correlation is $+0.463$.
After removing the fitted rate-related component from each architecture's
effect, the correlation is $+0.186$, indicating that approximately $60\%$
of the observed agreement is associated with the effective-rate component. The residual agreement remains positive in 145 of the 190 pairs and is
consistent with a contribution from the architecture-independent
near-freezing penalty described in Sec.~\ref{sec:freeze}. Accordingly, the
summary $(M_{\hat p},f)$ is more directly reusable than the original parameter grouping, whose effect also depends on
architecture-specific rate sensitivity (Sec.~\ref{sec:irreducible}).

\subsection{A One-Epoch Screen Is More Reliable for Ranking Than for Estimating Effects}
\label{sec:proxy}
Section~\ref{sec:tuned} clarifies what a short-budget measurement can
establish about the \emph{magnitude} of an effect: candidate ordering and
effect magnitude are distinct quantities. For each ablation architecture, the one-epoch screen ranked three strategies:
the best-performing strategy, a randomly selected strategy, and the
worst-performing strategy. All
three were then rerun for ten epochs (Fig.~\ref{fig:proxy}(a)). The ordering is preserved at epoch 10 in five of the six architectures
(mean $\rho=+0.75$; chance of $5/6$ or better $6.6\times10^{-4}$). The
best-to-worst gap grows from $9.1$ to $20.2$ points ($p=0.047$). By contrast,
the mean effect relative to the control shifts from $+1.9$ to $-2.6$ points,
and the effect changes sign in two of the six architectures. ConvNeXt is the architecture in which the ranking order reverses; its fitted
break-even level is among the three lowest of the six.
In this experiment, the one-epoch score is therefore more reliable for ranking
than for estimating effect size. The separation matters beyond this
study: proxy evaluations report rank correlation while the methods built on them
predict levels~\cite{lukasik2024evaluation}, and an ordering measured before a
rate decay need not survive it~\cite{wen2025fantastic}.

\subsection{Converting a Model-Zoo Measurement into a Retrained One}
\label{sec:zooconvert}
\begin{figure}[t]
\centering
\includegraphics[width=0.80\linewidth]{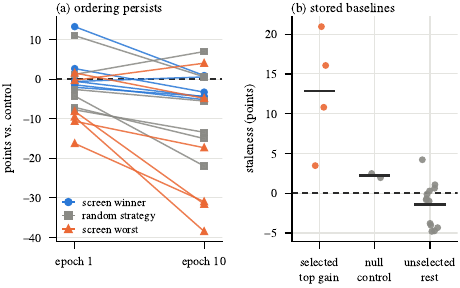}
\caption{(a)~Ordinal validity of the one-epoch screen: its best, random and
worst strategy against their shared control at epochs 1 and 10, six
architectures ($n=5$). (b)~Difference between a freshly trained control and the
stored accuracy, by the role assigned to each target before the re-check (group
means), which Eq.~(\ref{eq:shrink}) converts.}
\label{fig:proxy}
\end{figure}

Model zoos store an accuracy for each model. Reusing these stored values as
candidate baselines reduces the need to retrain reference models during
large-scale recipe search. Retraining fresh controls for 19 of the 25 held-out CIFAR-100 targets allows
the two baseline scales to be compared directly. The discrepancy is largest
for the selected targets: their stored accuracies are $12.8$ points below the
fresh controls in all four cases ($4/4$). The corresponding differences are
$+2.2$ points for null controls and $-1.4$ points for the 13 unselected targets
(one-sided Mann--Whitney $p=5.2\times10^{-4}$; Fig.~\ref{fig:proxy}(b)). Moreover, the larger the apparent gain
relative to the stored baseline, the larger this discrepancy
($\rho=+0.75$, $p=2.4\times10^{-4}$). An apparent $+9.0$-point gain becomes $-3.9$ against
fresh controls. Regressing one scale on the other,
\begin{equation}
\Delta_{\mathrm{fresh}} \;=\; 0.214\,\Delta_{\mathrm{stored}} \;-\; 0.019 ,
\label{eq:shrink}
\end{equation}
so approximately one-fifth of a gain measured against a model-zoo baseline
remains after the baseline is retrained ($r=0.43$). Tabular and surrogate
benchmarks often supply baseline accuracies for zero-cost and recipe-search
studies, and stored accuracies can be misleading when the models are
retrained~\cite{feng2025zcrobust}; agentic systems that report the best of many
short trials against a recipe not retuned for the target
setting~\cite{ning2026autoresearch} are one place where such a conversion may be
useful. The factor comes from a single protocol and should be recalibrated for
other platforms. Within the evaluated setting, stored baselines can be used
directly for the within-cluster analysis of Sec.~\ref{sec:crossds} and, after
conversion, for effect levels.

\section{A Protocol for Layerwise Learning-Rate Search}
\label{sec:protocol}
\label{sec:implications}
Table~\ref{tab:rules} summarizes the design recommendations and their evidence;
taken together, they define a lower-cost procedure than a conventional search.

\begin{table}[t]
\centering
\caption{The design rules this study supports, with the evidence for each.
Points are accuracy points; all tests are defined in Sec.~\ref{sec:design}.}
\label{tab:rules}
\scriptsize
\setlength{\tabcolsep}{3pt}
\renewcommand{\arraystretch}{1.0}
\begin{tabular}{p{0.46\linewidth}p{0.46\linewidth}}
\toprule
\textbf{Rule} & \textbf{Evidence}\\
\midrule
Rank candidates by $M_{\hat p}$, $\hat p\!=\!0.35$, not by the arithmetic
mean. & Zero-cost selection at the $10$th percentile rather than the $26$th; accuracy
regret of $1.12$ rather than $3.70$ points; effect-sign AUC of $0.86$.\\
Measure the rate sensitivity; do not predict it from the architecture. &
Every descriptor set scores a negative leave-one-architecture-out $R^2$.\\
Keep every group above $0.01\times$ the base rate. & $1.2$ points beyond the
effective-rate cost once the mean is curved ($t=-2.97$): a threshold, not a
gradient.\\
Decide the base rate first: a candidate's sign depends on it. & Tenfold rise in
$\eta_0$ lifts the pure rate cut in $6/6$ and ULMFiT from $2/6$ to $5/6$.\\
Evaluate grouping at a matched mean rate and validate its effect at long
budgets for each architecture. & $10/12$ cells at 10 epochs; $+6.4$ points at
100 epochs on ConvNeXt but $-0.88$ on GoogLeNet.\\
Assess the training budget before conducting a mean-rate search. & Decay as
$T^{-b}$,
$b=0.20$ to $0.57$; within the equivalence margin for $2/3$ tuned
baselines.\\
Transfer the summary rather than the grouping; adjust model-zoo gains before
interpreting their magnitude. & Approximately $60\%$ of the $+0.463$
cross-architecture agreement is associated with the effective-rate component; $\Delta_{\mathrm{fresh}}=0.214\,\Delta_{\mathrm{stored}}-0.019$.\\
\bottomrule
\end{tabular}
\end{table}

\section{Limitations}
\label{sec:limits}
The model is an empirical response model for from-scratch vision training
within the evaluated strategy space, rather than a general theory of layerwise
optimization. The exponent is estimated primarily on CIFAR-10: the positive
rate-response slope transfers across three datasets, but the exponent itself
has not yet been independently replicated. In addition, the evaluated
strategies only reduce or preserve group rates and use parameter-fraction
weighting in a strategy space dominated by depth-contiguous groups. Extensions
to boosted multipliers and alternative weighting schemes remain open.

The residual floor is identified through a post hoc decomposition of the same
trials used to estimate the exponent, rather than through an independent
threshold measurement. Its location is resolved only between $0.01$ and $0.1$,
reflecting the multiplier granularity of the evaluated strategy space.

The long-budget grouping evidence covers two architectures and one selected
strategy per architecture, with opposite responses. It therefore indicates
that a grouping-related contrast can persist, but does not establish a
universal sign or magnitude. The screening rate sweeps use one seed. The controlled identification
experiments focus on CIFAR-10 at 64\,px, whereas the cross-dataset analysis
primarily supports transfer of the rate-response slope. Conclusions about
practical equivalence are conditional on the prespecified $\pm1.53$-point
margin.

\section{Conclusion and Future Work}
For vision networks trained from scratch, the short-budget rate-related effects
of layerwise strategies can be summarized by an effective rate computed from the
multiplier profile. The relationship with training loss, the base-rate
intervention and the out-of-sample ULMFiT test support an optimization-progress
account of this channel. A concave power mean (exponent $0.35$,
cluster-bootstrap $95\%$ interval $[0.20,0.50]$) represents the aggregation
better than the arithmetic mean, accounting for $42.6\%$ rather than $36.4\%$ of
within-architecture variance, and improves training-free ranking of the top
candidate from the $26$th to the $10$th percentile.

The rate-related effect diminishes as the budget grows and lies within the
prespecified equivalence margin for two of the three tuned-baseline comparisons.
A grouping-related contrast at a matched arithmetic mean rate can persist at long
budgets, increasing on ConvNeXt and reversing on GoogLeNet, so it is
architecture-dependent rather than a uniform advantage of grouping. Because the
tested descriptors do not predict rate sensitivity for unseen architectures, a
small number of calibration runs remains useful even when candidate ranking is
training-free.

\textbf{Future work.} Priorities are to test whether the aggregation law
describes multipliers above one, to evaluate group weights that isolate small,
potentially high-leverage tensors, and to replicate the exponent on an
independent dataset with enough matched architecture clusters.

\section*{Acknowledgments}
\ifanon\else We thank our colleagues on the NNGPT/LEMUR project for the shared
evaluation infrastructure. \fi The AI system Claude (Anthropic) was used to generate the
charts in this article.

\bibliographystyle{IEEEtran}
\bibliography{references}

\begin{thebibliography}{10}
\providecommand{\url}[1]{#1}
\csname url@samestyle\endcsname
\providecommand{\newblock}{\relax}
\providecommand{\bibinfo}[2]{#2}
\providecommand{\BIBentrySTDinterwordspacing}{\spaceskip=0pt\relax}
\providecommand{\BIBentryALTinterwordstretchfactor}{4}
\providecommand{\BIBentryALTinterwordspacing}{\spaceskip=\fontdimen2\font plus
\BIBentryALTinterwordstretchfactor\fontdimen3\font minus \fontdimen4\font\relax}
\providecommand{\BIBforeignlanguage}[2]{{%
\expandafter\ifx\csname l@#1\endcsname\relax
\typeout{** WARNING: IEEEtran.bst: No hyphenation pattern has been}%
\typeout{** loaded for the language `#1'. Using the pattern for}%
\typeout{** the default language instead.}%
\else
\language=\csname l@#1\endcsname
\fi
#2}}
\providecommand{\BIBdecl}{\relax}
\BIBdecl

\bibitem{zoph2017nas}
B.~Zoph and Q.~V. Le, ``Neural architecture search with reinforcement learning,'' in \emph{Proc. Int. Conf. Learn. Represent. (ICLR)}, 2017.

\bibitem{liu2019darts}
H.~Liu, K.~Simonyan, and Y.~Yang, ``{DARTS}: Differentiable architecture search,'' in \emph{Proc. Int. Conf. Learn. Represent. (ICLR)}, 2019.

\bibitem{howard2018ulmfit}
J.~Howard and S.~Ruder, ``Universal language model fine-tuning for text classification,'' in \emph{Proc. 56th Annu. Meeting Assoc. Comput. Linguistics (ACL)}, 2018, pp. 328--339.

\bibitem{zhang2021revisiting}
T.~Zhang \emph{et~al.}, ``Revisiting few-sample {BERT} fine-tuning,'' in \emph{Proc. Int. Conf. Learn. Represent. (ICLR)}, 2021.

\bibitem{clark2020electra}
K.~Clark \emph{et~al.}, ``{ELECTRA}: Pre-training text encoders as discriminators rather than generators,'' in \emph{Proc. Int. Conf. Learn. Represent. (ICLR)}, 2020.

\bibitem{bao2022beit}
H.~Bao \emph{et~al.}, ``{BEiT}: {BERT} pre-training of image transformers,'' in \emph{Proc. Int. Conf. Learn. Represent. (ICLR)}, 2022.

\bibitem{you2017lars}
Y.~You, I.~Gitman, and B.~Ginsburg, ``Large batch training of convolutional networks,'' 2017, arXiv:1708.03888.

\bibitem{you2020lamb}
Y.~You \emph{et~al.}, ``Large batch optimization for deep learning: Training {BERT} in 76 minutes,'' in \emph{Proc. Int. Conf. Learn. Represent. (ICLR)}, 2020.

\bibitem{mlodozeniec2025completed}
B.~Mlodozeniec \emph{et~al.}, ``Completed hyperparameter transfer across modules, width, depth, batch and duration,'' 2025, arXiv:2512.22382.

\bibitem{zhou2026howtoset}
Y.~Zhou \emph{et~al.}, ``How to set the learning rate for large-scale pre-training?'' 2026, arXiv:2601.05049.

\bibitem{yosinski2014transferable}
J.~Yosinski \emph{et~al.}, ``How transferable are features in deep neural networks?'' in \emph{Proc. Adv. Neural Inf. Process. Syst. (NeurIPS)}, 2014, pp. 3320--3328.

\bibitem{kornblith2019better}
S.~Kornblith, J.~Shlens, and Q.~V. Le, ``Do better {ImageNet} models transfer better?'' in \emph{Proc. IEEE/CVF Conf. Comput. Vis. Pattern Recognit. (CVPR)}, 2019, pp. 2661--2671.

\bibitem{brock2017freezeout}
A.~Brock \emph{et~al.}, ``{FreezeOut}: Accelerate training by progressively freezing layers,'' 2017, arXiv:1706.04983.

\bibitem{zhou2023tempbalance}
Y.~Zhou \emph{et~al.}, ``Temperature balancing, layer-wise weight analysis, and neural network training,'' in \emph{Proc. Adv. Neural Inf. Process. Syst. (NeurIPS)}, 2023.

\bibitem{wang2025sharpness}
J.~Wang \emph{et~al.}, ``The sharpness disparity principle in transformers for accelerating language model pre-training,'' in \emph{Proc. Int. Conf. Mach. Learn. (ICML)}, 2025.

\bibitem{he2026onelr}
D.~He \emph{et~al.}, ``One {LR} doesn't fit all: Heavy-tail guided layerwise learning rates for {LLMs},'' in \emph{Proc. Int. Conf. Mach. Learn. (ICML)}, 2026.

\bibitem{pan2024lisa}
R.~Pan \emph{et~al.}, ``{LISA}: Layerwise importance sampling for memory-efficient large language model fine-tuning,'' in \emph{Proc. Adv. Neural Inf. Process. Syst. (NeurIPS)}, 2024.

\bibitem{zhang2025adammini}
Y.~Zhang \emph{et~al.}, ``{Adam-mini}: Use fewer learning rates to gain more,'' in \emph{Proc. Int. Conf. Learn. Represent. (ICLR)}, 2025.

\bibitem{hayou2024lorap}
S.~Hayou, N.~Ghosh, and B.~Yu, ``{LoRA}+: Efficient low rank adaptation of large models,'' in \emph{Proc. Int. Conf. Mach. Learn. (ICML)}, 2024.

\bibitem{yang2021tp5}
G.~Yang \emph{et~al.}, ``Tuning large neural networks via zero-shot hyperparameter transfer,'' in \emph{Proc. Adv. Neural Inf. Process. Syst. (NeurIPS)}, 2021.

\bibitem{bernstein2025modular}
J.~Bernstein and L.~Newhouse, ``Modular duality in deep learning,'' in \emph{Proc. Int. Conf. Mach. Learn. (ICML)}, 2025.

\bibitem{loshchilov2017sgdr}
I.~Loshchilov and F.~Hutter, ``{SGDR}: Stochastic gradient descent with warm restarts,'' in \emph{Proc. Int. Conf. Learn. Represent. (ICLR)}, 2017.

\bibitem{goyal2017accurate}
P.~Goyal \emph{et~al.}, ``Accurate, large minibatch {SGD}: Training {ImageNet} in 1 hour,'' 2017, arXiv:1706.02677.

\bibitem{pang2026balancing}
T.~Pang \emph{et~al.}, ``Balancing learning rates across layers: Exact two-step dynamics and optimal scaling in linear neural networks,'' in \emph{Proc. Int. Conf. Mach. Learn. (ICML)}, 2026.

\bibitem{kosson2024rotational}
A.~Kosson, B.~Messmer, and M.~Jaggi, ``Rotational equilibrium: How weight decay balances learning across neural networks,'' in \emph{Proc. Int. Conf. Mach. Learn. (ICML)}, 2024.

\bibitem{fan2025layerwise}
Z.~Fan \emph{et~al.}, ``Robust layerwise scaling rules by proper weight decay tuning,'' 2025, arXiv:2510.15262.

\bibitem{filatov2025norm}
O.~Filatov \emph{et~al.}, ``Optimal scaling needs optimal norm,'' 2025, arXiv:2510.03871.

\bibitem{kalra2026embedding}
D.~S. Kalra and M.~Barkeshli, ``Quantifying hyperparameter transfer and the importance of embedding layer learning rate,'' 2026, arXiv:2605.21486.

\bibitem{haas2024mup2}
M.~Haas \emph{et~al.}, ``$\mu$p$^2$: Effective sharpness aware minimization requires layerwise perturbation scaling,'' in \emph{Proc. Adv. Neural Inf. Process. Syst. (NeurIPS)}, 2024.

\bibitem{abdelfattah2021zerocost}
M.~S. Abdelfattah \emph{et~al.}, ``Zero-cost proxies for lightweight {NAS},'' in \emph{Proc. Int. Conf. Learn. Represent. (ICLR)}, 2021.

\bibitem{lukasik2024evaluation}
J.~Lukasik, M.~Moeller, and M.~Keuper, ``An evaluation of zero-cost proxies -- from neural architecture performance prediction to model robustness,'' \emph{Int. J. Comput. Vis.}, vol. 133, pp. 2635--2652, 2025.

\bibitem{feng2025zcrobust}
Y.~Feng \emph{et~al.}, ``Zero-cost proxy for adversarial robustness evaluation,'' in \emph{Proc. Int. Conf. Learn. Represent. (ICLR)}, 2025.

\bibitem{adhikari2026converg}
S.~P. Adhikari, R.~Timofte, and D.~Ignatov, ``Convergence theory for iterative {LLM}-based neural architecture search: A parametric cross-entropy framework with closed-form proxy reliability,'' 2026, arXiv:2605.30103.

\bibitem{lindauer2020best}
M.~Lindauer and F.~Hutter, ``Best practices for scientific research on neural architecture search,'' \emph{J. Mach. Learn. Res.}, vol.~21, no. 243, pp. 1--18, 2020.

\bibitem{li2019random}
L.~Li and A.~Talwalkar, ``Random search and reproducibility for neural architecture search,'' in \emph{Proc. Conf. Uncertainty Artif. Intell. (UAI)}, 2019, pp. 367--377.

\bibitem{white2023nas}
C.~White \emph{et~al.}, ``Neural architecture search: Insights from 1000 papers,'' 2023, arXiv:2301.08727.

\bibitem{lakens2017equivalence}
D.~Lakens, ``Equivalence tests: A practical primer for $t$ tests, correlations, and meta-analyses,'' \emph{Social Psychol. Personality Sci.}, vol.~8, no.~4, pp. 355--362, 2017.

\bibitem{romera2024funsearch}
B.~Romera-Paredes \emph{et~al.}, ``Mathematical discoveries from program search with large language models,'' \emph{Nature}, vol. 625, no. 7995, pp. 468--475, 2024.

\bibitem{zheng2023gpt4nas}
M.~Zheng \emph{et~al.}, ``Can {GPT-4} perform neural architecture search?'' 2023, arXiv:2304.10970.

\bibitem{chen2023evoprompting}
A.~Chen, D.~M. Dohan, and D.~R. So, ``{EvoPrompting}: Language models for code-level neural architecture search,'' in \emph{Proc. Adv. Neural Inf. Process. Syst. (NeurIPS)}, 2023.

\bibitem{jimenez2024swebench}
C.~E. Jimenez \emph{et~al.}, ``{SWE-bench}: Can language models resolve real-world {GitHub} issues?'' in \emph{Proc. Int. Conf. Learn. Represent. (ICLR)}, 2024.

\bibitem{yang2024sweagent}
J.~Yang \emph{et~al.}, ``{SWE-agent}: Agent-computer interfaces enable automated software engineering,'' in \emph{Proc. Adv. Neural Inf. Process. Syst. (NeurIPS)}, 2024.

\bibitem{ning2026autoresearch}
J.~Ning \emph{et~al.}, ``Auto research with specialist agents develops effective and non-trivial training recipes,'' 2026, arXiv:2605.05724.

\bibitem{kochnev2026nngpt}
R.~Kochnev \emph{et~al.}, ``{NNGPT}: Rethinking {AutoML} with large language models,'' in \emph{Proc. IEEE/CVF Conf. Comput. Vis. Pattern Recognit. Workshops (CVPRW)}, 2026, pp. 3262--3272.

\bibitem{goodarzi2025lemur}
A.~T. Goodarzi \emph{et~al.}, ``{LEMUR} neural network dataset: Towards seamless {AutoML},'' 2025, arXiv:2504.10552.

\bibitem{mateen2026lrsched}
H.~Mateen, R.~Timofte, and D.~Ignatov, ``Systematic evaluation of learning rate scheduling strategies across heterogeneous architectures,'' 2026, arXiv:2607.08511.

\bibitem{everett2024scaling}
K.~E. Everett \emph{et~al.}, ``Scaling exponents across parameterizations and optimizers,'' in \emph{Proc. Int. Conf. Mach. Learn. (ICML)}, 2024.

\bibitem{liu2026elr}
Z.~Liu \emph{et~al.}, ``Effective learning rate governs loss dynamics in language model pretraining,'' 2026, arXiv:2608.24814.

\bibitem{hui2024qwen25coder}
B.~Hui \emph{et~al.}, ``{Qwen2.5-Coder} technical report,'' 2024, arXiv:2409.12186.

\bibitem{hu2022lora}
E.~J. Hu \emph{et~al.}, ``{LoRA}: Low-rank adaptation of large language models,'' in \emph{Proc. Int. Conf. Learn. Represent. (ICLR)}, 2022.

\bibitem{krizhevsky2009cifar}
A.~Krizhevsky, ``Learning multiple layers of features from tiny images,'' Univ. Toronto, Toronto, ON, Canada, Tech. Rep., 2009.

\bibitem{brown2001interval}
L.~D. Brown, T.~T. Cai, and A.~DasGupta, ``Interval estimation for a binomial proportion,'' \emph{Statist. Sci.}, vol.~16, no.~2, pp. 101--133, 2001.

\bibitem{good2005permutation}
P.~Good, \emph{Permutation, Parametric, and Bootstrap Tests of Hypotheses}, 3rd~ed.\hskip 1em plus 0.5em minus 0.4em\relax New York, NY, USA: Springer, 2005.

\bibitem{benjamini1995fdr}
Y.~Benjamini and Y.~Hochberg, ``Controlling the false discovery rate: A practical and powerful approach to multiple testing,'' \emph{J. Roy. Statist. Soc. B, Statist. Methodol.}, vol.~57, no.~1, pp. 289--300, 1995.

\bibitem{bates2015lme4}
D.~Bates \emph{et~al.}, ``Fitting linear mixed-effects models using {lme4},'' \emph{J. Statist. Softw.}, vol.~67, no.~1, pp. 1--48, 2015.

\bibitem{wu2026seesaw}
A.~Meterez \emph{et~al.}, ``Seesaw: Accelerating training by balancing batch size and learning rate scheduling,'' in \emph{Proc. Int. Conf. Learn. Represent. (ICLR)}, 2026.

\bibitem{shulgin2026hpscaling}
E.~Shulgin \emph{et~al.}, ``Deriving hyperparameter scaling laws via modern optimization theory,'' 2026, arXiv:2603.15958.

\bibitem{dahl2023benchmarking}
G.~E. Dahl \emph{et~al.}, ``Benchmarking neural network training algorithms,'' 2023, arXiv:2306.07179.

\bibitem{kasimbeg2025algoperf}
P.~Kasimbeg \emph{et~al.}, ``Accelerating neural network training: An analysis of the {AlgoPerf} competition,'' in \emph{Proc. Int. Conf. Learn. Represent. (ICLR)}, 2025.

\bibitem{wen2025fantastic}
K.~Wen \emph{et~al.}, ``Fantastic pretraining optimizers and where to find them,'' 2025, arXiv:2509.02046.

\bibitem{shen2026looped}
J.~Shen and X.~Li, ``Deeper backpropagation or more updates? training looped language models across budgets,'' 2026, arXiv:2610.38422.

\bibitem{eimer2023hyperparameters}
T.~Eimer, M.~Lindauer, and R.~Raileanu, ``Hyperparameters in reinforcement learning and how to tune them,'' in \emph{Proc. Int. Conf. Mach. Learn. (ICML)}, 2023.

\end{thebibliography}

\end{document}